\documentclass[11pt]{article}
\usepackage{paclic33}
\usepackage{times}
\usepackage{latexsym}
\usepackage{amsmath}
\usepackage{multirow}
\usepackage{url}

\usepackage{tikz,subfigure}
\usetikzlibrary{positioning}

\usepackage{xcolor}
\usepackage{amssymb}
\usepackage{mathtools}
\usepackage{setspace}
\usepackage{here}
\usepackage{proof, lscape}
\usepackage{ebproof}
\usepackage{natbib}
\usepackage{gb4e}

\newcommand{\NKC}{\textsc{Comp}}
\newcommand{\leaf}[2]{\infer{#1}{\mbox{#2}}}
\newcommand{\LF}[1]{\ensuremath{\mathbf{#1}}}
\newcommand{\semtree}[2]{\begin{tabular}{cc}\ensuremath{#1} : \\ \ensuremath{#2}\end{tabular}}
\newcommand{\syncatsr}[2]{\ensuremath{#1} : \ensuremath{#2}}
\newcommand{\fun}[2]{\ensuremath{\mathbf{#1}({#2})}}
\newcommand{\funt}[3]{\ensuremath{\mathbf{#1}({#2},{#3})}}
\newcommand{\funtall}[2]{\ensuremath{\funt{tall}{#1}{#2}}}
\newcommand{\funadjer}[3]{\ensuremath{\exists \delta \,(\funt{#1}{#2}{\delta} \wedge \neg \,\funt{#1}{#3}{\delta})}}
\newcommand{\funtaller}[2]{\ensuremath{\funadjer{tall}{#1}{#2}}}
\newcommand{\thadj}[1]{\ensuremath{\theta_{#1}}}
\newcommand{\thtall}{\ensuremath{\thadj{\text{tall}}}}
\newcommand{\thsho}{\ensuremath{\thadj{\text{short}}}}
\newcommand{\funasadj}[3]{\ensuremath{\forall \delta(\funt{#1}{#2}{\delta} \rightarrow \funt{#1}{#3}{\delta})}}
\newcommand{\funtallp}[1]{\ensuremath{\mathbf{tall}({#1}, \thtall)}}
\newcommand{\funshortp}[1]{\ensuremath{\mathbf{short}({#1}, \thsho)}}

\newcommand{\funasadje}[3]{\ensuremath{\forall e(\funt{#1}{#2}{e} \rightarrow \funt{#1}{#3}{e})}}

\newcommand{\funcl}[2]{\ensuremath{\fun{\thadj{\textit{#1}}}{#2}}}

\newcommand{\ccgiv}{\ensuremath{S\backslash \mathit{NP}}}
\newcommand{\ccgqp}{\ensuremath{S/(S\backslash \mathit{NP})}}
\newcommand{\ccgraise}{\ensuremath{\mathit{NP}^{\uparrow}}}
\newcommand{\ccgivs}{\ensuremath{\ccgiv/(\ccgiv)}}
\newcommand{\ccgdegl}{\ensuremath{\ccgiv \backslash D}}

\newcommand{\ccgcl}{\ensuremath{S/S}}
\newcommand{\ccghl}{\ensuremath{\ccgiv/\mathit{AP}}}

\newcommand{\ccgerc}{\ensuremath{\ccgiv/\ccgraise\backslash\mathit{AP}}}
\newcommand{\ccgerca}{\ensuremath{\ccgiv/\ccgraise/\mathit{AP}}}
\newcommand{\ccger}{\ensuremath{\ccgiv/(S \backslash D) \backslash \mathit{AP}}}
\newcommand{\ccgderc}{\ensuremath{\ccgiv/\ccgraise \backslash D \backslash \mathit{AP}}}
\newcommand{\ccgerd}{\ensuremath{\ccgiv/D \backslash \mathit{AP}}}

\newcommand{\ccgdth}{\ensuremath{\ccgiv \backslash (\ccgiv/\ccgraise)/\ccgraise}}

\newcommand{\ccgercc}{\ensuremath{\ccgiv/(\ccgqp) \backslash (\ccgdegl)}}
\newcommand{\ccgadjer}{\ensuremath{\ccgiv/(\ccgqp)}}
\newcommand{\ccgpos}{\ensuremath{\ccgiv/(\ccgdegl)}}
\newcommand{\ccgmov}{\ensuremath{\ccgiv/\ccgraise \backslash (\ccgiv/\mathit{NP})/N}}
\newcommand{\ccgell}{\ensuremath{\ccgiv/\ccgraise\backslash (\ccgiv/\mathit{NP})/N/AP}}

\newcommand{\rulelabelsize}{\scriptsize}
\newcommand{\FA}{\mbox{\rulelabelsize $<$}}
\newcommand{\BA}{\mbox{\rulelabelsize $>$}}

\newcommand{\SR}[1]{\begin{tabular}{c}$#1$\end{tabular}}

\newcommand{\lamadj}[1]{\ensuremath{\lambda \delta x.\mathbf{#1}(x, \delta)}}
\newcommand{\lamnp}[1]{\ensuremath{\lambda P. P({#1})}}
\newcommand{\lampos}[1]{\ensuremath{\lambda A. A(\theta_{#1})}}

\newcommand{\lamadjd}[2]{\ensuremath{\lambda x.\mathbf{#1}(x, {#2})}}
\newcommand{\lamer}{\ensuremath{\lambda A K x. \exists \delta (A(\delta)(x) \wedge \neg K(\delta))}}
\newcommand{\lamerc}{\ensuremath{\lambda A Q x. \exists \delta (A(\delta)(x) \wedge \neg Q(A(\delta)))}}
\newcommand{\lamerd}{\ensuremath{\lambda A \delta' x. \exists \delta(A(\delta)(x) \wedge (\delta > \delta'))}}
\newcommand{\lamderc}{\ensuremath{\lambda A \delta' Q x. \forall \delta(Q(A(\delta)) \rightarrow A(\delta+\delta')(x))}}
\newcommand{\lamadjerc}[2]{\ensuremath{\lambda Q x. \exists \delta ({#1} \wedge \neg Q({#2}))}}

\newcommand{\lamas}{\ensuremath{\lambda A Q x. \forall \delta (Q(A(\delta)) \rightarrow A(\delta)(x))}}
\newcommand{\lamtallerx}[2]{\ensuremath{\lambda x. \funtaller{#1}{#2}}}
\newcommand{\lamth}{\ensuremath{\lambda Q W x. Q(\lambda y. W(\lambda P. P(y))(x))}}

\newcommand{\lammov}{\ensuremath{\lambda N G Q z. \exists \delta(\exists x(N(x) \wedge G(\lambda P. P(x))(z) \wedge \funt{many}{x}{\delta}})}
\newcommand{\lammovv}{\ensuremath{ \wedge \neg \exists y(N(y) \wedge Q(G(\lambda P. P(y))) \wedge \funt{many}{y}{\delta}))}}
\newcommand{\lamell}{\ensuremath{\lambda A N G Q z. \exists \delta(\exists x(N(x) \wedge G(\lambda P. P(x))(z) \wedge A(\delta)(x)))}}
\newcommand{\lamhas}{\ensuremath{ \wedge \neg \exists y(N(y) \wedge Q(G(\lambda P. P(y))) \wedge A(\delta)(x))}}
\newcommand{\lamis}{\ensuremath{ \wedge \neg Q(\lambda y.(N(y) \wedge A(\delta)(x)))}}
\newcommand{\mea}[4]{\ensuremath{\exists \delta(\funt{#1}{\LF{#2}}{\delta} \wedge (\delta #3 #4))}}

\newcommand{\RightApp}{\mbox{\scriptsize$>$}}
\newcommand{\LeftApp}{\mbox{\scriptsize$<$}}
\newcommand{\RightB}{\mbox{\scriptsize$>\!\mathbf{B}$}}

\newcommand{\RightT}{\mbox{\scriptsize$>\!\mathbf{T}$}}

\newcommand{\lex}{\RightT}

\newcommand{\fplus}{\ensuremath{F^+}}
\newcommand{\fminus}{\ensuremath{F^-}}

\newcommand{\adjneg}[3]{\ensuremath{\forall e \forall x(\textbf{#1}(x, e) \leftrightarrow \forall \delta ((\delta \,{#2}\, e) \rightarrow \neg \,\textbf{#3}(x, \delta)))}}
\newcommand{\adjto}[3]{\ensuremath{\forall e \forall x(\textbf{#1}(x, e) \leftrightarrow \forall \delta ((\delta \,{#2}\, e) \rightarrow \textbf{#3}(x, \delta)))}}
\newcommand{\negadj}[3]{\ensuremath{\forall e \forall x(\neg \,\textbf{#1}(x, e) \leftrightarrow \forall \delta ((\delta \,{#2}\, e) \rightarrow \textbf{#3}(x, \delta)))}}
\newcommand{\thres}[2]{\ensuremath{\theta_{\text{#1}} > \theta_{\text{#2}}}}

\newcommand{\ToE}{\mbox{\scriptsize$(\to\!\!\mathbf{E})$}}

\newcommand{\AllE}{\mbox{\scriptsize$(\forall \mathbf{E})$}}

\newcommand{\cpstar}{\mbox{\scriptsize$(\mathbf{CP}\star)$}}

\title{A CCG-based Compositional Semantics and Inference System \\ for Comparatives}
  
\author{Izumi Haruta \\ Ochanomizu University \\  \texttt{haruta.izumi@is.ocha.ac.jp}
        \AND
        Koji Mineshima \\ Ochanomizu University \\ \texttt{mineshima.koji@ocha.ac.jp} \And
        Daisuke Bekki \\ Ochanomizu University \\ \texttt{bekki@is.ocha.ac.jp}}

\date{}

\begin{document}
\maketitle

\begin{abstract}
Comparative constructions play an important role in natural language inference.
However, attempts to study semantic representations and
logical inferences for comparatives 
from the computational perspective
are not well developed,
due to the complexity of their syntactic structures and inference patterns.
In this study, using a framework based on Combinatory Categorial Grammar (CCG),
we present a compositional semantics that maps various comparative constructions in English
to semantic representations and introduces an inference system that effectively handles
logical inference with comparatives, including those involving numeral adjectives, 
antonyms, and quantification.
We evaluate the performance of our system on the FraCaS test suite and
show that the system can handle a variety of complex logical inferences with comparatives.
\end{abstract}

\section{Introduction}
\label{sec:introduction}

Gradability is a pervasive phenomenon in natural language and plays an important role in natural language understanding.
Gradable expressions can be characterized
in terms of the notion of \textit{degree}.
Consider the following examples:
\begin{exe}
    \ex 
    \begin{xlist}
        \ex My car is more \textit{expensive} than yours. \label{car}
        \ex My car is \textit{expensive}. \label{pos}
    \end{xlist}
\end{exe}
The sentence (\ref{car}), in which the comparative form of the gradable adjective \textit{expensive} is used, compares the price of two cars,
making it a comparison between degrees.
The sentence (\ref{pos}), which contains the positive form of the adjective, can be regarded as a construction that compares the price of the car to some 
implicitly given degree (i.e., price).

In formal semantics, 
many in-depth analyses use a semantics of gradable expressions that relies on the notion of degree~
\citep[][among others]{cresswell1976semantics,kennedy97,heim2000degree,lassiter2017graded}.
Despite this, 
meaning representations
and inferences for gradable expressions have not been well developed from the perspective of computational semantics in previous research~\citep{pulman2007formal}.
Indeed,
a number of logic-based inference systems
have been proposed for the task of Recognizing Textual Entailment (RTE),
a task
to determine whether a set of premises entails a given hypothesis~\citep{bos2008wide,maccartney2008modeling,mineshima2015higher,abzianidze2016natural,bernardy2017type}.
However, these logic-based systems have performed relatively poorly
on inferences with gradable constructions,
such as those collected in the FraCaS test suite~\citep{cooper1994fracas},
a standard benchmark dataset for evaluating logic-based RTE systems (see \S\ref{sec:eval} for details).

There are at least two obstacles to developing a comprehensive computational analysis of gradable constructions.
First, the syntax of gradable constructions is diverse, as shown in (\ref{tayousei}):
 \begin{exe}
 \ex  \label{tayousei}
 \begin{xlist}
 \ex \label{posi}
 Ann is tall.\hfill{(Positive)} 
 \ex 
 Ann is taller than Bob. \hfill{(Phrasal)}
 \ex 
 Ann is taller than Bob is.
 \hfill{(Clausal)} 
 \ex Ann is as tall as Bob.
 \hfill{(Equative)}
 \ex Ann is $2^{\prime\prime}$ taller than Bob. \hfill{(Differential)}\label{2tall}
 \end{xlist}
 \end{exe}
 
\noindent
In the examples above,
(\ref{tayousei}c) is a clausal comparative in which \textit{tall} is missing from the subordinate \textit{than}-clause.
(\ref{tayousei}e) is an example of a differential comparative in which a measure phrase, 
$2^{\prime\prime}$ (\textit{2 inches}), appears.
The diversity of syntactic structures makes it difficult to 
provide a compositional semantics for comparatives in a computational setting.

Second, gradable constructions give rise to
various inference patterns that require
logically complicated steps.
For instance, consider (\ref{fig:map1}):
\begin{exe}
\ex \label{fig:map1}
    {\renewcommand\arraystretch{1.3}
    \raisebox{-0.6cm}{
  	\begin{tabular}{cl}
    	$P_1$: & Mary is taller than 4 feet.\\
        $P_2$: & Harry is shorter than 4 feet.\\
  	\hline
    	$H$: & Mary is taller than Harry.\\
  	\end{tabular}}
  	}
\end{exe}
To logically derive $H$ from $P_1$ and $P_2$, one has to assign the proper meaning representations
to each sentence,
and those representations 
include numeral expressions (\textit{4 feet}), antonyms (\textit{short/tall}), and their interaction with comparative constructions.

For these reasons, gradable constructions pose an important challenge to logic-based approaches to RTE,
serving as a testbed to act as a bridge between formal semantics and computational semantics.

In this paper, we provide (i) a compositional semantics to map various gradable constructions in English to semantic representations (SRs) and (ii) an inference system that derives logical inference with gradable constructions in an effective way.
We will mainly focus on gradable adjectives and their comparative forms
as representatives of gradable expressions,
leaving the treatment of other gradable
constructions such as verbs and adverbs to future work.

We use Combinatory Categorial Grammar (CCG)~\citep{Steedman2000} as a syntactic component of
our system and the so-called \textit{A-not-A analysis}~\citep{seuren1973comparative,klein1980semantics,klein1982interpretation,schwarzschild2008semantics} to
provide semantic representations
for comparatives (\S\ref{sec:background}, \S3).
We use ccg2lambda \citep{ccg2lambda:2016} to 
implement compositional semantics
to map CCG derivation trees to SRs.
We introduce an axiomatic system \NKC\
for inferences with comparatives
in typed logic with equality and
arithmetic operations (\S4).
We use a state-of-the-art prover
to implement the \NKC\ system.
We evaluate our system\footnote{
All code is available at:\\
https://github.com/izumi-h/fracas-comparatives\_adjectives}
on the two sections of the FraCaS test suite (\textsc{adjective} and \textsc{comparative})
and show that it can handle various complex inferences with gradable adjectives and comparatives.

\begin{table*}[t]
  \caption{Semantic representations of basic comparative constructions}
  \centering
  \scalebox{0.9}{$
  {\renewcommand\arraystretch{1.1}
  \begin{tabular}{lll}\hline
    \textbf{Type} &
    \textbf{Example} &
    \textbf{SR}\\
    \hline
    Increasing Comparatives & Mary is taller than Harry. & \funtaller{\LF{m}}{\LF{h}}\\
    Decreasing Comparatives & Mary is less tall than Harry. & $\exists \delta(\neg\funtall{\LF{m}}{\delta} \wedge \funtall{\LF{h}}{\delta})$\\
    Equatives & Mary is as tall as Harry. & \funasadj{tall}{\LF{h}}{\LF{m}}\\
    \hline
    \end{tabular}
    }
    $}
    \label{fig:comparatives1}
\end{table*}

\begin{table*}[t]
\caption{Semantic representations of complex comparative constructions}

  \centering
  \scalebox{0.9}{$\displaystyle
  \begin{tabular}{lll}\hline
    \textbf{Type} &
    \textbf{Example} &
    \textbf{SR}\\
    \hline
    Subdeletion Comparatives & Mary is taller than the bed is long. & $\exists \delta(\funtall{\LF{m}}{\delta} \wedge \neg \,\funt{long}{\LF{the(bed)}}{\delta})$\\
    Measure phrase comparatives & Mary is taller than 4 feet. & \mea{tall}{m}{>}{4^\prime}\\
    Differential Comparatives & Mary is 2 inches taller than Harry. & $\forall \delta(\funtall{\LF{h}}{\delta} \rightarrow \funtall{\LF{m}}{\delta+2''})$\\
    Negative Adjectives & Mary is shorter than Harry. & \funadjer{short}{\LF{m}}{\LF{h}}\\
    \hline
    \end{tabular}
    $}
    \label{fig:comparatives}
\end{table*}

\section{Background}
\label{sec:background}

\subsection{
Comparatives in degree-based semantics}
\label{ssec:semantics}

To analyze gradable adjectives, we use the two-place
predicate of entities and degrees as developed in
degree-based semantics~\citep{klein1982interpretation,kennedy97,heim2000degree,schwarzschild2008semantics}.
For instance, the sentence \textit{Ann is 6 feet tall}
is analyzed as \funtall{\LF{Ann}}{\text{6 feet}},
where \funtall{x}{\delta} is read as ``$x$ is (at least) as tall as degree $\delta$''.\footnote{
For simplicity, we do not consider the internal structure of a measure phrase like \textit{6 feet}.
For an explanation of why 
\funtall{x}{\delta} is not treated as ``$x$ is \textit{exactly} as tall as $\delta$'', see, e.g., \cite{klein1982interpretation}.}

In degree-based semantics,
there are at least two types of analyses for comparatives.
Consider (\ref{basic}), a schematic example for a comparative construction.
\begin{exe}
\ex $A$ is taller than $B$ is. \label{basic}
\end{exe}

\noindent
The first approach is based on the maximality operator \citep{stechow1984comparing,heim2000degree}.
Using the maximality operator ($\max$) as illustrated in (\ref{max}), 
the sentence (\ref{basic}) is analyzed as a statement asserting that the maximum degree $\delta_1$ of $A$'s tallness
is greater than the maximum degree $\delta_2$ of $B$'s tallness.
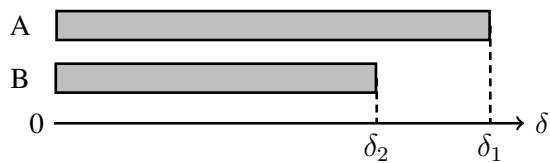
\begin{exe}
\ex $\max (\lambda \delta. \funtall{A}{\delta}) > \max (\lambda \delta. \funtall{B}{\delta})$
\label{max}
\end{exe}
\begin{tikzpicture}
    \node(nA) at (0,0) {A};
    \node(nB) at (0,-0.7) {B};
	\node[fill=black!25, rectangle, draw=black, line width=1pt, text width=5.5cm,text centered,minimum height=1em, right=0.5em of nA] (A) {};
	\node[fill=black!25, rectangle, draw=black, line width=1pt, text width=4cm,text centered,minimum height=1em, right=0.5em of nB] (B) {};
	\coordinate (O) at (0.45,-1.3) node at (O) [left] {0};
	\coordinate (P) at (6.7,-1.3);
	\draw [->, line width=1pt] (O)--(P) node at (P) [right] {$\delta$};
	\draw [densely dashed, line width=1pt] (6.25,0)--(6.25,-1.3) node at (6.25,-1.3) [below] {\large $\delta_1$};
	\draw [densely dashed, line width=1pt] (4.74,-0.7)--(4.74,-1.3) node at (4.74,-1.3) [below] {\large $\delta_2$};
\end{tikzpicture}

The other approach is the A-not-A analysis~\citep{seuren1973comparative,klein1980semantics,klein1982interpretation,schwarzschild2008semantics}.
In this type of analysis, (\ref{basic}) is treated as stating
that there exists a degree $\delta^\prime$ of tallness that $A$ satisfies
but $B$ does not, as shown in (\ref{anota}).

\begin{exe}
\ex $\funtaller{A}{B}$ \label{anota}
\end{exe}

\begin{tikzpicture}
    \node(nA) at (0,0) {A};
    \node(nB) at (0,-0.7) {B};
	\node[fill=black!25, rectangle, draw=black, line width=1pt, text width=5.5cm,text centered,minimum height=1em, right=0.5em of nA] (A) {};
	\node[fill=black!25, rectangle, draw=black, line width=1pt, text width=4cm,text centered,minimum height=1em, right=0.5em of nB] (B) {};
	\coordinate (O) at (0.45,-1.3) node at (O) [left] {0};
	\coordinate (P) at (6.7,-1.3);
	\draw [->, line width=1pt] (O)--(P) node at (P) [right] {$\delta$};
	\draw [densely dashed, line width=1pt] (6.25,0)--(6.25,-1.3) node at (6.25,-1.3) [below] {\large $\delta_1$};
	\draw [densely dashed, line width=1pt] (4.74,-0.7)--(4.74,-1.3) node at (4.74,-1.3) [below] {\large $\delta_2$};
	\draw [densely dashed, line width=1pt] (5.3,0.2)--(5.3,-1.3) node at (5.3,-1.3) [below] {\large $\delta^\prime$};
\end{tikzpicture}

Although the two analyses are related as illustrated in the figures (\ref{max}) and (\ref{anota}),
we can say that the
A-not-A analysis is less complicated and easier to handle
than the maximality-based analysis from a computational perspective,
mainly because it only involves constructions in first-order logic (FOL).\footnote{See \cite{van2008comparatives} for
a more detailed comparison of the two approaches.}
We thus adopt the A-not-A analysis 
and extend it to various types of comparative constructions
for which inference is efficient in our system.

\subsection{Basic syntactic assumptions}
\label{ssec:syntax}

There are two approaches to the syntactic
analysis of comparative constructions.
The first is the \textit{ellipsis} approach \citep[e.g.][]{kennedy97},
in which 
phrasal comparatives such as (\ref{tayousei}b),
are derived from the corresponding
clausal comparatives, such as (\ref{tayousei}c).
The other is the \textit{direct} approach~\citep[e.g.][]{hendriks95},
which treats phrasal and clausal comparatives
independently and does not derive one from the other.
An argument against the ellipsis approach
is that it has difficulties
in accounting for coordination such as that in (\ref{ex:petra})~\citep{hendriks95}.
\begin{exe}
	\ex	 \label{ex:petra}
	\begin{xlist}
		\ex Someone at the party drank more vodka than wine. \label{drank}
		\ex Someone at the party drank more vodka than someone at the party drank wine. \label{someone}
	\end{xlist}
\end{exe}
Here, (\ref{ex:petra}a), a phrasal comparative
with an existential NP \textit{someone},
does not have the same meaning as
the corresponding clausal comparative
(\ref{ex:petra}b); 
the person who drank vodka and the one who drank wine do not have to be the same person in (\ref{ex:petra}b), whereas they must be the same person in (\ref{ex:petra}a).\footnote{See \citet{hendriks95} and
\citet{kubota2015against} for other arguments against the ellipsis approach.}
In this study, we adopt the direct approach
and use CCG 
to formalize the syntactic component of our system.

\section{Framework}
\label{sec:sr}

\subsection{Semantic representations}
\label{ssec:AnotA}

Table \ref{fig:comparatives1} shows the SRs for basic constructions of comparatives
under the A-not-A analysis we adopt.
Using this standard analysis, we also provide
SRs for more complex constructions,
including subdeletion, measure phrases,
and negative adjectives. 
Table \ref{fig:comparatives} summarizes the SRs
for these constructions.

Some remarks are in order about
how our system handles various linguistic
phenomena related to gradable adjectives and comparatives.

\paragraph{Antonym and negative adjectives}

\textit{Short} is the antonym of \textit{tall}, which is represented as \funt{short}{x}{\delta} and has the meaning ``the height of $x$ is less than or equal to $\delta$''.
Thus, we distinguish between 
the monotonicity property of
positive adjectives such as \textit{tall} and \textit{fast} and that of
negative adjectives such as \textit{short} and \textit{slow}.
For positive adjectives,
if \funt{tall}{x}{\delta} is true,
then $x$ satisfies all heights below $\delta$;
by contrast, for negative adjectives,
if \funt{short}{x}{\delta} is true,
then $x$ satisfies all the heights above $\delta$.

In general, for a positive adjective $\fplus$
and a negative adjective $\fminus$,
(\ref{tallshort}a) and (\ref{tallshort}b) hold, respectively.

\begin{exe}
\ex \label{tallshort}
$\forall \delta_1 \forall \delta_2: \delta_1 > \delta_2 \to$  
    \begin{xlist}
    \ex $\forall x (\fplus(x, \delta_1) \rightarrow \fplus(x, \delta_2))$
    \ex $\forall x(\fminus(x, \delta_2) \rightarrow  \fminus(x, \delta_1))$
    \end{xlist}
\end{exe}

\paragraph{Positive form and comparison class}

As mentioned in \S\ref{sec:introduction},
the positive form of an adjective
is regarded as involving
comparison to some threshold that can be inferred from the context of the utterance. 
We write \funcl{F}{A}
to denote the contextually specified threshold
for a predicate $F$ given a set $A$,
which is called
\textsc{comparison class}~\citep{klein1982interpretation}.
When a comparison class is implicit, as in 
(\ref{postall}a) and (\ref{posshort}a),
we use the universal set $U$
as a default comparison class\footnote{
In this case, we do not consider the context-sensitivity of the implicit comparison class.
See \cite{narisawa2013204} 
for work on this topic
in computational linguistics.};
we typically abbreviate \funcl{F}{U}
as \thadj{F}.
Thus, (\ref{postall}a) is represented
as (\ref{postall}b), which
means that the height of Mary is more than or equal to the threshold $\thtall$.
Similarly,
the SR of (\ref{posshort}a) is (\ref{posshort}b),
which means that the height of Mary is
less than or equal to the threshold $\thsho$.
\begin{exe}
    \ex \label{postall}
    \begin{xlist}
    \ex Mary is tall. \label{tall}
    \ex $\funtallp{\LF{m}}$ \label{tallp}
    \end{xlist}
    \ex \label{posshort}
    \begin{xlist}
    \ex Mary is short. \label{short}
    \ex \small 
    $\funshortp{\LF{m}}$ \label{shortp}
    \end{xlist}
\end{exe}
A threshold can be explicitly constrained by an NP modified by a gradable adjective. 
Thus,
(\ref{ex:excompclass}a)
can be interpreted as (\ref{ex:excompclass}b),
relative to an explicit comparison class,
namely, the sets of animals.\footnote{
Here and henceforth, when an example appears in the FraCaS dataset,
we refer to the ID of the sentence in the dataset.}

\begin{exe}
    \ex \label{ex:excompclass}
        \begin{xlist}
            \ex Mickey is a small animal. 
            \hfill{\small (FraCaS-204)}    
            \ex 
{\small$\funt{small}{\LF{m}}{\fun{\thadj{small}}{\LF{animal}}} \wedge \fun{animal}{\LF{m}}$}
        \end{xlist}
\end{exe}

\paragraph{Numerical adjectives}

We represent a numerical adjective such as \textit{ten} in \textit{ten orders} 
by the predicate \funt{many}{x}{n},
with the meaning that the cardinality of $x$ is at least $n$, where $n$ is a positive integer~\citep{hackl2000comparative}.
For example, \textit{ten orders} is analyzed
as $\lambda x. (\LF{order}(x) \wedge \funt{many}{x}{\text{10}})$.
The following shows the SRs of some typical sentences involving numerical adjectives.
\begin{exe}
\ex 
    \begin{xlist}
    \ex Mary won ten orders.
    \ex $\exists x(\LF{order}(x) \wedge
    \funt{won}{\LF{m}}{x} \\\wedge \funt{many}{x}{10})$
    \end{xlist}
\ex 
    \begin{xlist}
    \ex Mary won many orders.
    \ex $\exists \delta \exists x (\LF{order}(x)
    \wedge \funt{won}{\LF{m}}{x}
    \\\wedge 
    \funt{many}{x}{\delta} \wedge (\theta_{\text{many}}(\LF{order}) < \delta))$
    \end{xlist}
\ex 
    \begin{xlist}
    \ex Mary won more orders than Harry.  \label{moreor}
    \ex $\exists \delta (\exists x (\LF{order}(x) \wedge \funt{won}{\LF{m}}{x} \\\wedge \funt{many}{x}{\delta}) \wedge \neg 
    \exists y (\LF{order}(y) \\\wedge \funt{won}{\LF{h}}{y} \wedge \funt{many}{y}{\delta}))$
    \end{xlist}
\end{exe}

\begin{table*}[t]
\caption{Lexical entries in CCG-style compositional semantics}
\centering
\scalebox{0.8}{$\displaystyle
  \begin{tabular}{lll}
  \hline
    PF  & CCG categories & SR  \\
    \hline 
    tall  & $AP$ & \lamadj{tall}\\
    Mary & $\mathit{NP}$ & \LF{mary}\\
    is & \ccgivs & $id$\\
    $4^{\prime}$ & $D$ & $4'$\\
    than$_{\text{simp}}$ & \ccgcl & $id$\\
    than$_{\text{deg}}$ & $D/D$ & $id$\\
    than$_{\text{gq}}$ & \ccgdth & \lamth\\
    pos & \ccghl & \lampos{A}\\
    -er$_{\text{simp}}$ & \ccgerc & \lamerc\\ 
    -er$_{\text{sub}}$ & \ccger & \lamer\\ 
    -er$_{\text{mea}}$ & \ccgerd & \lamerd\\ 
    -er$_{\text{diff}}$ & \ccgderc & \lamderc\\
    as$_{\text{simp}}$ & \ccgerca & \lamas\\
    as$_{\text{cl}}$ & \ccgcl &  $id$\\
    more$_{\text{num}}$ & \ccgmov & \lammov \\ 
    & & \lammovv\\
    more$_{\text{is}}$ & \ccgell & \lamell\\
    & & \lamis\\
    more$_{\text{has}}$ & \ccgell & \lamell\\ 
    & & \lamhas\\
    \hline
  \end{tabular}
$}

\label{fig:entry}
\vspace{-10pt}
\end{table*}

\begin{figure*}[t]
\vspace{1em}
\centering
   \scalebox{0.8}{
\infer[\RightApp]{\syncatsr{S}{\funtaller{\text{\LF{m}}}{\LF{h}}}}{
    \infer[\lex]{\semtree{\ccgqp}{\lamnp{\LF{m}}}}{
     \leaf{\syncatsr{NP}{\LF{m}}}{Mary}
    }
    &
    \infer[\RightApp]{\semtree{\ccgiv}{\lamtallerx{x}{\LF{h}}}}{
    \leaf{\semtree{\ccgivs}{id}}{is}
    &\hspace{-1cm}
    \infer[\RightApp]{\semtree{\ccgiv}{\lamtallerx{x}{\LF{h}}}}{
    \infer[\LeftApp]{\semtree{\ccgadjer}{\lamadjerc{\funtall{x}{\delta}}{\lamadjd{tall}{\delta}}}}{
     \leaf{\semtree{\ccgdegl}{\lamadj{tall}}}{tall}
    &
    \leaf{\semtree{\ccgercc}{\lamerc}}{-er_{\text{simp}}}
    }
    &
     \infer[\RightB]{\semtree{\ccgqp}{\lamnp{\LF{h}}}}{
     \leaf{\semtree{\ccgcl}{id}}{than_{\text{simp}}}
    &
    \infer[\lex]{\semtree{\ccgqp}{\lamnp{\LF{h}}}}{
    \leaf{\syncatsr{NP}{\LF{h}}}{Harry}
    }
    }
    }
        }
        }
        }
    \caption{Derivation tree of \textit{Mary is taller than Harry}}
    \label{eq:p1}
\end{figure*}

\begin{figure}[t]
\centering
\scalebox{0.53}{
\infer[\RightApp]{\semtree{S}{\funtallp{\LF{h}}}}{
    \infer[\lex]{\semtree{\ccgqp}{\lamnp{\LF{h}}}}{
     \leaf{\semtree{NP}{\LF{h}}}{Harry}
    }
    &
    \infer[\RightApp]{\semtree{\ccgiv}{\lamadjd{tall}{\thtall}}}{
    \leaf{\semtree{\ccgivs}{id}}{is}
    &
    \infer[\RightApp]{\semtree{\ccgiv}{\lamadjd{tall}{\thtall}}}{
    \leaf{\semtree{\ccgpos}{\lampos{A}}}{pos}
    &
    \leaf{\semtree{\ccgdegl}{\lamadj{tall}}}{tall}
    }}
}
}
    \caption{Derivation tree of \textit{Harry is tall}}
    \label{eq:p2}
\end{figure}

\subsection{Compositional semantics in CCG}
\label{ssec:mapping}

Here we give an overview of how to compositionally derive the SRs for comparative constructions
in the framework of CCG~\citep{Steedman2000}.
In the CCG-style compositional semantics,
each lexical item is assigned both a syntactic category and an SR (represented as a $\lambda$-term).
In this study, we newly introduce 
the syntactic category $D$ for degree
and assign \ccgdegl\ to gradable adjectives.
For instance, the adjective \textit{tall}
has the category \ccgdegl\
and the corresponding SR is $\lambda \delta. \lambda x. \LF{tall}(x,\delta)$.

Table \ref{fig:entry} lists the
lexical entries for representative lexical items used in the proposed system.
We abbreviate the CCG category \ccgdegl\ for adjectives
as $AP$ and \ccgqp\ (a type-raised NP) as \ccgraise.\footnote{
We also abbreviate $\lambda X_1. \ldots \lambda X_n. M$ as $\lambda X_1 \ldots X_n. M$.}

The suffix \textit{-er} for comparatives such as \textit{taller} is categorized into four types:
clausal and phrasal comparatives (-er$_{\text{simp}}$),
subdeletion comparatives (\mbox{-er}$_{\text{sub}}$),
measure phrase comparatives (-er$_{\text{mea}}$),
and differential comparatives (-er$_{\text{diff}}$).
We assume that equatives are constructed from as$_{\text{simp}}$ and as$_{\text{cl}}$;
for instance, the equative sentence in Table \ref{fig:comparatives1}
corresponds to \textit{Mary is as$_{\text{simp}}$ tall as$_{\text{cl}}$ Harry}.
For measure phrase comparatives, such as \textit{Mary is taller than 4 feet}, we use than$_{\text{deg}}$;
and for comparatives with numerals, such as (\ref{moreor}),
we use more$_{\text{simp}}$.

On the basis of these lexical entries,
we can compositionally map various comparative constructions to  suitable SRs.
Some example derivation trees for comparative constructions
are shown in Figure \ref{eq:p1} and \ref{eq:p2}.
An advantage of using CCG as a syntactic theory
is that the \textit{function composition} rule
$\mathbf{(\RightB)}$ can be used for phrasal comparatives
such as that in Figure \ref{eq:p1},
where the VP \textit{is tall} is missing from the subordinate \textit{than}-clause.
For positive forms, we use the empty element \textit{pos}
of category $S\backslash NP/(S\backslash NP\backslash D)$,
as shown in Figure \ref{eq:p2}.\footnote{
Note that the role played by the empty element \textit{pos} here
can be replaced by imposing a unary type-shift rule
from $S\backslash NP\backslash D$ to
$S\backslash NP$.}

\begin{figure*}[t]
\centering
\scalebox{0.58}{
\deduce{\SR{\forall y(\fun{person}{y} \to \exists \delta(\funt{tall}{\LF{m}}{\delta} \,\wedge\, \neg \funt{tall}{y}{\delta}))}}{
\infer[\BA]{S :}{
\deduce{\SR{\lambda P.P(\LF{m})}}{
\infer[\lex]{\ccgqp :}{
\deduce{\SR{\LF{m}}}{
\infer{NP :}{\mbox{Mary}}}}
} & 
\deduce{\SR{\lambda x.\forall y(\fun{person}{y} \to \exists \delta(\funt{tall}{x}{\delta} \,\wedge\, \neg \funt{tall}{y}{\delta}))}}{
\infer[\BA]{S\backslash NP :}{
\deduce{\SR{id}}{
\infer{S\backslash NP/(S\backslash NP) :}{\mbox{is}}}
& 
\deduce{\SR{\lambda x.\forall y(\fun{person}{y} \to \exists \delta(\funt{tall}{x}{\delta} \,\wedge\, \neg \funt{tall}{y}{\delta}))}}{
\infer[\FA]{S\backslash NP :}{
\deduce{\SR{\lambda Q x.\exists \delta(\funt{tall}{x}{\delta} \,\wedge\, \neg Q(\lambda x.\funt{tall}{x}{\delta}))
}}{
\infer[\FA]{S\backslash NP/(S/(S\backslash NP)) :}{
\deduce{\SR{\lambda \delta x.\funt{tall}{x}{\delta}}}{
\infer{S\backslash NP\backslash D :}{\mbox{tall}}} & 
\deduce{\SR{\lamerc}}{
\infer{S\backslash NP/(S/(S\backslash NP))\backslash (S\backslash NP\backslash D) :}{\mbox{-er_{\text{simp}}}}}}} & 
\deduce{\SR{\lambda W x.\forall y(\fun{person}{y} \to
W(\lambda P. P(y))(x)
)}}{
\infer[\BA]{S\backslash NP\backslash (S\backslash NP/(S/(S\backslash NP))) :}{
\deduce{\SR{\lamth
}}{
\infer{S\backslash NP\backslash (S\backslash NP/(S/(S\backslash NP)))/(S/(S\backslash NP)) :}{\mbox{than_{\text{gq}}}}} & 
\deduce{\SR{\lambda P.\forall y(\fun{person}{
y} \to P(y))}}{
\infer{\ccgqp :}{\mbox{everyone}}}
}}}}}}}}
}
    \caption{Derivation tree of \textit{Mary is taller than everyone}}
    \label{eve}
\end{figure*}

\paragraph{Quantification}
When determiners such as \textit{all} or \textit{some} appear in \textit{than}-clauses, we need to consider the scope of the corresponding quantifiers \citep{larson1988scope}.
As examples, (\ref{ex:quant1}a) and (\ref{ex:quant2}a)
are assigned the SRs in (\ref{ex:quant1}b) and (\ref{ex:quant2}b),
respectively.
 
\begin{exe}
\ex \label{ex:quant1}
    \begin{xlist}
    \ex Mary is taller than everyone. \label{ex:all}
    \ex $\forall y(\fun{person}{y} \\\rightarrow \funtaller{\LF{m}}{y})$ \label{ex:alls}
    \end{xlist}
\ex \label{ex:quant2}
    \begin{xlist}
    \ex Mary is taller than someone. \label{ex:some}
    \ex $\exists y(\fun{person}{y} \\\wedge \funtaller{\LF{m}}{y})$
    \end{xlist}
\end{exe}

\noindent
Figure \ref{eve} shows a derivation tree for (\ref{ex:quant1}a).
Here, \textit{everyone} in \textit{than}-clause takes
scope over the degree quantification in the main clause.
For this purpose, we use the lexical entry for than$_{\text{gq}}$ in Table \ref{fig:entry}, which handles these cases of generalized
quantifiers.

\paragraph{Conjunction and disjunction}
Conjunction (\textit{and}) and disjunction (\textit{or}) 
appearing in a \textit{than}-clause
show different behaviors in scope taking, as pointed out by \cite{larson1988scope}.
For instance, in (\ref{ex:and}a), the conjunction \textit{and} takes wide scope over the main clause, whereas in (\ref{ex:or}a),
the disjunction \textit{or} can take narrow scope;
thus, we can infer \textit{Mary is taller than Harry} from both (\ref{ex:and}a) and (\ref{ex:or}a).
These readings are
represented as in (\ref{ex:and}b) and (\ref{ex:or}b), respectively.

\begin{exe}
\ex \label{ex:and}
    \begin{xlist}
    \ex Mary is taller than Harry and Bob. 
    \ex $\funtaller{\LF{m}}{\LF{h}} \\
    ~~~~~~~\wedge \funtaller{\LF{m}}{\LF{b}}$
    \end{xlist}
\ex \label{ex:or}
    \begin{xlist}
    \ex Mary is taller than Harry or Bob. \label{orsent}
    \ex $\exists \delta(\funt{tall}{\LF{m}}{\delta} \\\wedge \neg (\funt{tall}{\LF{h}}{\delta} \vee \funt{tall}{\LF{b}}{\delta}))$
    \end{xlist}
\end{exe}
The difference in scope for these sentences can be derived by using
than$_{\text{simp}}$ and than$_{\text{gq}}$:
than$_{\text{simp}}$ derives the narrow-scope reading
(cf. the derivation tree in Figure \ref{eq:p1})
and than$_{\text{gq}}$ derives the wide-scope reading
(cf. the derivation tree in Figure \ref{eve}).

\paragraph{Attributive comparatives}
The sentence \textit{APCOM has a more important customer than ITEL} (FraCaS-244/245) can have two interpretations, i.e., (\ref{ellis}a) and (\ref{ellhas}a), where the difference is in the verb of the \textit{than}-clause.

\begin{exe}
    \ex \label{ellis}
    \begin{xlist}
        \ex APCOM has a more important customer than ITEL \underline{is}. \hfill{(FraCaS-244)}
        \ex \small $\exists \delta(\exists x(\fun{customer}{x} \\\wedge \funt{has}{\LF{a}}{x} \wedge \funt{important}{x}{\delta}) \\\wedge \neg(\fun{customer}{\LF{i}} \wedge \funt{important}{\LF{i}}{\delta}))$ \label{is}
    \end{xlist}
    \ex \label{ellhas}
    \begin{xlist}
        \ex APCOM has a more important customer than ITEL \underline{has}. \hfill{(FraCaS-245)}
        \ex \small $\exists \delta(\exists x(\fun{customer}{x} \wedge \funt{has}{\LF{a}}{x} \\\wedge \funt{important}{x}{\delta}) \\\wedge \neg\exists y(\fun{customer}{y} \wedge \funt{has}{\LF{i}}{y} \\\wedge \funt{important}{y}{\delta}))$ \label{has}
    \end{xlist}
\end{exe}
We use more$_{\text{is}}$ or more$_{\text{has}}$ in Table \ref{fig:entry} to give the compositional derivations of 
the SRs in (\ref{is}) and (\ref{has}), respectively.

\begin{table*}[t]
 \caption{Axioms of \NKC}
\small
\centering
{\renewcommand\arraystretch{1.2}
  \begin{tabular}{l|l}
  \hline
$\mathbf{(TH)}$ & \thres{\fplus}{\fminus}\\
$\mathbf{(CP)}$ & $\forall x \forall y (\funadjer{\textit{F}}{x}{y} \rightarrow (\forall e (\funt{\textit{F}}{y}{e} \rightarrow \funt{\textit{F}}{x}{e})))$\\
\hline\hline
$\mathbf{(Ax_1)}$ & \adjto{\fminus}{\geq}{\fminus}\\
$\mathbf{(Ax_2)}$ & \adjto{\fplus}{\leq}{\fplus}\\
\hline\hline
$\mathbf{(Ax_3)}$ & \adjneg{\fminus}{>}{\fplus}\\
$\mathbf{(Ax_4)}$ & \adjneg{\fplus}{<}{\fminus}\\
$\mathbf{(Ax_5)}$ & \negadj{\fminus}{\leq}{\fplus}\\
$\mathbf{(Ax_6)}$ & \negadj{\fplus}{\geq}{\fminus}\\
\hline
  \end{tabular}
 }
\label{fig:nkc1}
\end{table*}

\section{Inferences with comparatives}
\label{sec:inference}

We introduce an inference system \NKC\ for logical reasoning with gradable adjectives and comparatives
based on the SRs under the A-not-A analysis presented in \S\ref{sec:sr}.
Table \ref{fig:nkc1} lists some axioms of \NKC\ for inferences with comparatives.
Here, $F$ is an arbitrary gradable predicate, $F^+$ a positive adjective, and $F^-$ a negative adjective.\footnote{
We also use an axiom for privative adjectives such as \textit{former}, drawn from \cite{mineshima2015higher}.}

$\mathbf{(CP)}$ is the so-called Consistency Postulate~\citep{klein1982interpretation},
an axiom asserting that if there is a degree satisfied by $x$ but not by $y$, then every degree satisfied by $y$ is satisfied by $x$ as well.
By $\mathbf{(CP)}$, we can derive the following inference rule.
\begin{center}
    \ebproofnewstyle{small}{
    rule margin = .5ex,
    template = $\inserttext$ }
    \begin{prooftree}[small]
        \hypo{ \funadjer{\mathbf{F}}{x}{y} }
        \infer[left label = \cpstar]1[]{ \forall e (\funt{\mathbf{F}}{y}{e} \!\rightarrow\! \funt{\mathbf{F}}{x}{e}) }
    \end{prooftree}
\end{center}
Using this rule, the inference from \textit{Mary is taller than Harry} and \textit{Harry is tall} to \textit{Mary is tall} can be derived as shown in Figure \ref{fig:reasoning}.

\begin{figure}[H]
    \centering
    \ebproofnewstyle{small}{
        separation = .5em, rule margin = .5ex,
        template = $\inserttext$ }
    \begin{prooftree}[small]
        \hypo{ \funtaller{\LF{m}}{\LF{h}} }
        \infer[left label= \cpstar]1[]
        { \funasadje{tall}{\LF{h}}{\LF{m}} }
        \infer[left label= \AllE]1[]
        { \funtallp{\LF{h}}\!\to\!\funtallp{\LF{m}} }
        \hypo{ \funtallp{\LF{h}} }
        \infer[left label =  \ToE]2[]{ \funtallp{\LF{m}} }
    \end{prooftree}
    \caption{Example of a proof}
    \label{fig:reasoning}
\end{figure}

$\mathbf{(Ax_1)}$ and $\mathbf{(Ax_2)}$ are axioms for positive and negative adjectives described in (\ref{tallshort}).
The axioms from $\mathbf{(Ax_3)}$ to $\mathbf{(Ax_6)}$ formalize the entailment relations between antonym predicates.
For instance, the inference of (\ref{fig:map1}) mentioned in \S\ref{sec:introduction}
is first mapped to the following SRs.
\begin{exe}
\ex
\raisebox{-0.6cm}{
{\tabcolsep = 5pt
    {\renewcommand\arraystretch{1.3}
  	\begin{tabular}{clcl}
    	{\small $P_1$:} & \mea{tall}{m}{>}{4^\prime}\\
    	{\small $P_2$:} & \mea{short}{h}{<}{4^\prime}\\
  	\hline
    	{\small $H$:} & \funtaller{\LF{m}}{\LF{h}}\\
  	\end{tabular}
	}
	}
}
\end{exe}
Then, it can be easily shown that $H$ follows from $P_1$ and $P_2$, using the axioms
$\mathbf{(Ax_2)}$ and $\mathbf{(Ax_3)}$.

\begin{table*}[t]
\caption{Accuracy on FraCaS test suite.
`\#All' shows the number of all problems 
and `\#Single' the number of single-premise problems.
}
 \centering
\scalebox{0.95}{$\displaystyle
        {\renewcommand\arraystretch{1.1}
	\begin{tabular}{c||c|cccccc}
	\hline
        Section & \#All & Ours & B\&C & Nut & MINE & LP & M\&M (\#Single)\\
        \cline{3-4}
        \hline\hline
        \textsc{adjectives} & 22 & \textbf{1.00} & .95 & .32 & .68 & .73 & .80* (15)\\
        \textsc{comparatives} & 31 & \textbf{.94} & .56 & .45 & .48 & - & .81* (16)\\
        \hline
  	\end{tabular}
  	}
  	$}
  	
  	\label{fig:fracas2}
\end{table*}

\section{Implementation and evaluation}
\label{sec:eval}

To implement a full inference pipeline, one needs three components:
(a) a syntactic parser that maps input sentences to
CCG derivation trees,
(b) a semantic parser that maps CCG derivation trees
to SRs, and 
(c) a theorem prover that proves entailment relations between these SRs.
In this study, we use manually constructed CCG trees as inputs
and implement components (b) and (c).\footnote{
CCG parsers for English, such as C\&C parser~\citep{Clark2007} based on CCGBank~\citep{HockenmaierSteedman},
are widely used, but there is a gap between the outputs of these existing parsers and
the syntactic structures we assume for the analysis of comparative constructions as described in \S\ref{sec:sr}.
We leave a detailed comparison between those structures to another occasion.
We also have to leave the task of combining our system with off-the-shelf CCG parsers for future research.
}
For component (b), 
we use ccg2lambda\footnote{
https://github.com/mynlp/ccg2lambda
}
as a semantic parser and implement a set of templates
corresponding to the lexical entries in Table \ref{fig:entry}.
The system takes a CCG derivation tree as an input
and outputs a logical formula as an SR.
For component (c), we use the off-the-shelf theorem prover \textit{Vampire}\footnote{
https://github.com/vprover/vampire
}
and implement the set of axioms described in \S\ref{sec:inference}.

Suppose that the logical formulas corresponding to given premise sentences
are $P_1, \ldots, P_n$ and that the logical formula corresponding to the hypothesis (conclusion) is $H$.
Then, 
the system outputs \textit{Yes} if $P_1 \wedge \cdots \wedge P_n \to H$ can be proved by a theorem prover,
and outputs \textit{No} if the negation of the hypothesis  (i.e., $P_1 \wedge \cdots P_n \to \neg H$) 
can be proved.
If both of them fail, it tries to construct a counter model; if a counter model is found,
the system outputs \textit{Unknown}.
Since the main purpose of this implementation is to test the correctness of our semantic analysis and inference system,
the system returns \textit{error} if a counter model is not constructed with the size of an allowable model restricted.

We evaluate our system on the FraCaS test suite.
The test suite is a collection of semantically complex inferences for various linguistic phenomena drawn
from the literature on formal semantics and is categorized into nine sections.
Out of the nine sections, we use \textsc{adjectives} (22 problems) and \textsc{comparatives} (31 problems).
The distribution of gold answers is: (yes, no, unknown) = (9, 6, 7) for \textsc{adjectives} and (19, 9, 3) for \textsc{comparatives}.
Table \ref{fig:fracas} lists some examples.

\begin{table}[h]
\small
\caption{Examples of entailment problems from the FraCaS test suite}
\smallskip
 \centering
\scalebox{0.95}{$\displaystyle
    {\renewcommand\arraystretch{1.3}
  	\begin{tabular}{l|l}\hline
  	    \multicolumn{2}{l}{FraCaS-198 (\textsc{adjectives}) \ Answer: No} \\
    	\hline
    	\textbf{Premise 1} & John is a former university student.\\
    	\textbf{Hypothesis} & John is a university student.\\
    	\hline\hline
    	\multicolumn{2}{l}{FraCaS-224 (\textsc{comparatives}) \ Answer: Yes}  \\
    	\hline
    	\textbf{Premise 1} & The PC-6082 is as fast as the ITEL-XZ.\\
    	\textbf{Premise 2} & The ITEL-XZ is fast.\\
    	\textbf{Hypothesis} & The PC-6082 is fast.\\
    	\hline\hline
    	\multicolumn{2}{l}{FraCaS-229 (\textsc{comparatives}) \ Answer: No} \\
    	\hline
    	\textbf{Premise 1} & The PC-6082 is as fast as the ITEL-XZ.\\
    	\textbf{Hypothesis} & The PC-6082 is slower than the ITEL-XZ.\\
    	\hline\hline
    	\multicolumn{2}{l}{FraCaS-231 (\textsc{comparatives}) \ Answer: Unknown} \\
    	\hline
    	\textbf{Premise 1} & ITEL won more orders than APCOM did.\\
    	\textbf{Hypothesis} & APCOM won some orders.\\
    	\hline\hline
    	\multicolumn{2}{l}{FraCaS-235 (\textsc{comparatives}) \ Answer: Yes} \\
    	\hline
    	\textbf{Premise 1} & ITEL won more orders than APCOM.\\
    	\textbf{Premise 2} & APCOM won ten orders.\\
    	\textbf{Hypothesis} & ITEL won at least eleven orders.\\
    	\hline
  	\end{tabular}
  	}
  	$}
  	\label{fig:fracas}
\end{table}

Table \ref{fig:fracas2} gives the results of the evaluation.
We compared our system with existing logic-based RTE systems. 
B\&C~\citep{bernardy2017type} is an RTE-system based on Grammatical Framework~\citep{ranta2011grammatical} and 
uses the proof assistant Coq for theorem proving.
The theorem proving part is not automated
but manually checked.
Nut~\citep{bos2008wide} and MINE~\citep{mineshima2015higher} use a CCG parser~\citep[C\&C parser;][]{Clark2007} and implement a theorem-prover for RTE based on FOL and higher-order logic, respectively.
LP~\citep{abzianidze2016natural}
is a system, LangPro,
that uses two CCG parsers (C\&C parser and EasyCCG;
\citep{lewis2014ccg}) and implements
a tableau-based natural logic inference system.
M\&M~\citep{maccartney2008modeling}
uses an inference system for natural logic based on monotonicity calculus.
M\&M was only evaluated for a subset
of the FraCaS test suite, considering
single-premise inferences and excluding multiple-premise
inferences.
These four systems, Nut, MINE, LP, and M\&M,
are fully automated.

Although direct comparison is impossible
due to differences in automation and
the set of problems used for evaluation (single-premise or multiple-premise),
our system achieved a considerable improvement
in terms of accuracy.
It should be noted that by using arithmetic implemented
in Vampire our system correctly performed complex inferences from numeral expressions such as that in FraCaS-235 
(see Table \ref{fig:fracas}).
Because we did not implement a syntactic parser and used gold CCG trees instead,
the results show the upper bound
of the logical capacity of our system.
Note also that the five systems (B\&C, MINE, LP, M\&M, and ours) were developed in part to solve inference problems in FraCaS, where there is no separate test data for evaluation. Still, these problems are linguistically very challenging; from a linguistic perspective, the point of evaluation is to see \textit{how} each system can solve a given inference problem.
Overall, the results of evaluation suggest that a semantic parser
based on degree semantics
can, in combination with a theorem prover, achieve
high accuracy for a range of complex inferences with
adjectives and comparatives.

There are two problems in the \textsc{comparatives}
section that our system did not solve:
the inference from $P$ to $H_1$
and the one from $P$ to $H_2$, both having the gold answer \textit{Yes}.
\begin{center} \small
\begin{tabular}{l} 
$P$: ITEL won more orders than the APCOM contract.\\
$H_1$: ITEL won the APCOM contract.
(FraCaS-236)\\
$H_2$: ITEL won more than one order.
(FraCaS-237)
\end{tabular}
\end{center}
To solve these inferences in a principled way,
we will need to consider a more systematic way of
handling comparative constructions
that expects at least two patterns with missing verb phrases.
\section{Conclusion}
\label{sec:conclusion}

We proposed a CCG-based compositional semantics for gradable adjectives and comparatives
using the A-not-A analysis studied in formal semantics.
We implemented a system that maps
CCG trees to suitable SRs and performs
theorem proving for RTE.
Our system achieved high accuracy on the sections for adjectives and comparatives in FraCaS.

In future work, we will further extend the empirical coverage of our system. 
In particular, we will cover deletion operations like Gapping in comparatives, as well as gradable expressions other than adjectives.
Combining our system with a CCG parser is also left for future work.

\paragraph{Acknowledgement}
This work was supported by JSPS KAKENHI Grant Number JP18H03284.
\bibliographystyle{apalike}
\bibliography{main}
\end{document}